\documentclass[12pt,letterpaper]{article}
\usepackage[margin=1in]{geometry}
\usepackage{hyperref}       
\usepackage{url}            
\usepackage{booktabs}       
\usepackage{amsfonts}       
\usepackage{nicefrac}       
\usepackage{microtype}      
\usepackage{graphicx}
\usepackage{natbib}
\usepackage{titling}
\PassOptionsToPackage{numbers, compress}{natbib}
\usepackage{amsmath}
\usepackage{mathtools}

\usepackage{color}
\usepackage{subcaption}
\usepackage{multirow}
\usepackage{caption}

\author{Sinead A. Williamson$ ^{1,2} $, Michael Minyi Zhang$ ^{3} $, Paul Damien$ ^{1,2} $}
\title{A New Class of Time Dependent Latent Factor Models with Applications}
\date{\texttt{sinead.williamson@mccombs.utexas.edu, mz8@cs.princeton.edu, paul.damien@mccombs.utexas.edu}\\
	$^{1}$ Department of Statistics and Data Science. The University of Texas at Austin.\\
	$^{2}$ Department of Information, Risk, and Operations Management. McCombs School of Business. The University of Texas at Austin.\\
	$^{3}$ Department of Computer Science. Princeton University.\\
	\vspace{2em} 
	\today}	
\begin{document}
	\maketitle	
	\begin{abstract} 
 In many applications, observed data are influenced by some combination of latent causes. For example, suppose sensors are placed inside a building to record responses such as temperature, humidity, power consumption and noise levels. These random, observed responses are typically affected by many unobserved, latent factors (or features) within the building such as the number of individuals, the turning on and off of electrical devices, power surges, etc. These latent factors are usually present for a contiguous period of time before disappearing; further, multiple factors could be present at a time. This paper develops new probabilistic methodology and inference methods for random object generation influenced by latent features exhibiting temporal persistence. Every datum is associated with subsets of a potentially infinite number of hidden, persistent features that account for temporal dynamics in an observation. The ensuing class of dynamic models constructed by adapting the Indian Buffet Process --- a probability measure on the space of random, unbounded binary matrices --- finds use in a variety of applications arising in operations, signal processing, biomedicine, marketing, image analysis, etc. Illustrations using synthetic and real data are provided.
	\end{abstract} 
\section{Introduction}
Random object generation is a broad topic, since the word ``object'' has many connotations in mathematics and applied probability. For example, ``object'' could refer to a matrix or a polynomial. Indeed, observed data are random objects; for instance, a vector of observables in a regression context satisfies transparently the idea of a probabilistic “object” \citep{Leemis:2006}. Of late, a class of random object models is growing in popularity, namely Latent Factor (or Feature) Models, abbreviated LFM. The theory and use of these models lie at the intersection of probability theory, Bayesian inference, and simulation methods, particularly Markov chain Monte Carlo (MCMC). Saving the formal description of LFMs for future sections, consider the following heuristics of certain key ideas central to the paper.

Latent variables are unobserved, or are not directly measurable. Parenting skill, speech impediments, socio-economic status, and quality of life are some examples of these. Latent variables could also correspond to a ``true'' variable observed with error. Examples would include iron intake measured by a food frequency, self-reported weight, and lung capacity measured by forced expiratory volume in one second. In Bayesian hierarchical modeling, latent variables are often used to represent unobserved properties or hidden causes of data that are being modeled \citep{Bishop:1998}. Often, these variables have a natural interpretation in terms of certain underlying but unobserved features of the data; as examples, thematic topics in a document or motifs in an image. The simplest of such models, which we will refer to as Latent Variable Models (LVMs), typically use a finite number of latent variables, with each datum related to a single latent variable \citep{Bishop:1998,McLaughlan:2000}. This class of models includes finite mixture models, where a datum is associated with a single latent mixture component, and Hidden Markov Models (HMMs) where each point in a time series is associated with a single latent state \citep{Baum:Petrie:1996}. All data associated with a given latent parameter are assumed to be independently and identically simulated according to a distribution parametrized by that latent parameter.

Greater flexibility can be obtained by allowing multiple latent features for each datum. This allows different aspects of a datum to be shared with different subsets of the dataset. For example, two articles may share the theme ``science'', but the second article may also exhibit the theme ``finance''. Similarly, a picture of a dog in front of a tree has aspects in common with both  pictures of trees and pictures of dogs. Models that allow multiple features are typically referred to as Latent Factor Models (LFMs). Examples of LFMs include Bayesian Principle Component Analysis where data are represented using a weighted superposition of latent factors, and Latent Dirichlet Allocation where data are represented using a mixture of latent factors; see \citet{Roweis:Ghahramani:1999} for a review of both LVMs and LFMs.

In the majority of LVMs and LFMs, the number of latent variables is finite and pre-specified. The appropriate cardinality is often hard to determine \emph{a priori} and, in many cases, we do not expect our training set to contain exemplars of all possible latent variables. These difficulties have led to the increasing popularity of LVMs and LFMs where the number of latent variables associated with each datum or object is potentially unbounded; see,  \citep{Antoniak:1974,Teh:Jordan:Beal:Blei:2006, Griffiths:Ghahramani:2005,Titsias:2007,Broderick:Mackay:Paisley:Jordan:2015}. These latter probabilistic models with an infinite number of parameters are referred to as nonparametric latent variable models (npLVMs) and nonparametric latent factor models (npLFMs). These models generally tend to provide richer inferences than their finite-dimensional counterparts, since deeper relationships between the unobserved variables and the observed data could be obtained by relaxing finite distributional assumptions about the probability generating mechanism.

In many applications, data are assumed \emph{exchangeable}, in that no information is conveyed by the order in which data are observed. Even though exchangeability is a weaker (hence preferable) assumption than independent and identically distributed data, often times, observed data are time-stamped emissions from some evolving process. That is, the ordering (or dependency) is crucial to understanding the entire random data-generating mechanism. There are two types of dependent data that, typically, arise in practice. It is convenient to use terminology from the biomedical literature to distinguish the two. \emph{Longitudinal dependency} refers to situations where one records multiple entries from the same random process over a  period of time. In AIDS research, a biomarker such as a CD4 lymphocyte cell count is observed intermittently for a patient and its relation to time of death is of interest. In a different context, the ordering of frames in a video sequence or the ordering of windowed audio spectra in a piece of music within a time interval are crucial to our understanding of the entire video or musical piece. 

\emph{Epidemiological dependency} corresponds to situations where our data generating mechanism involves multiple random processes, but where we typically observe each single process at only one covariate value; that is, that is, single records from multiple entities constitute the observed data. For instance, in an annual survey on diabetic indicators one might interview a different group of people each year; the observations correspond to different random processes (i.e. different individuals), but still capture global trends. Or consider articles published in a newspaper: while the articles published today are distinct from those published yesterday, there are likely to be general trends in the themes covered over time. 

Most research using npLFMs has focused on the exchangeable setting with non-dependent nonparametric LFMs being deployed in a number of application areas  \citep{Wood:Griffiths:Ghahramani:2006,Ruiz:Valera:Blanco:PerezCruz:2014,Meeds:Ghahramani:Neal:Roweis:2007}.  A number of papers have developed npLFMs for epidemiological dependence \citep{Foti:Futoma:Rockmore:Williamson:2013,Ren:Wang:Carin:Dunson:2011,Zhou:Yang:Sapiro:Dunson:Carin:2011,Rao:Teh:2009}. In these settings we are often able to make use of conjugacy to develop reasonably efficient stochastic simulation schemes. In addition, several nonparametric priors for LFMs have been proposed for longitudinally dependent data ~\citep{Williamson:Orbanz:Ghahramani:2010,Gershman:Frazier:Blei:2015}, but unfortunately these papers, by virtue of their modeling approaches, require computationally complex inference protocols. Furthermore, these existing epidemiologically and longitudinally dependent methods are often invariant under time reversal. This is often a poor choice for modeling temporal dynamics, where the direction of causality means that they dynamics are not invariant under time reversal.  


In this paper, we introduce a new class of npLFMs that is suitable for time (or longitudinally) dependent data. From a modeling perspective, the focus is on npLFMs rather than npLVMs since the separability assumptions underlying LVMs are overly restrictive for most real data. Specifically, we follow the tradition of generative or simulation-based npLFMs. A Bayesian approach is natural in this framework since the form of npLFMs needed to better model temporal dependency involves the use of probability distributions on function spaces; the latter idea is commonly referred to as Bayesian nonparametric inference \citep{Walker:Damien:Laud:Smith:1999}.

The primary aims of this research are the following. First, to develop a class of npLFMs with practically useful attributes to generate random objects in a variety of applications. These attributes include an unbounded number of latent factors;  capturing temporal dynamics in the data; and the tracking of \emph{persistent} factors over time. The significance of this class of models is best described with a simple, yet meaningful, example. Consider a flautist playing a musical piece. At very short time intervals if the flautist is playing a B$\flat$ at time $t$, it is likely that note would still be playing at time $t+1$. Arguably, this is a continuation of a single note instance that begins at time $t$ and persists to time $t+1$ (or beyond). Unlike current approaches, our proposed time-dynamic model captures this (persistent latent factor) dependency in the musical notes from time $t$ to $t+1$ (or beyond). The second goal of this research is to develop a general Markov chain Monte Carlo algorithm to enable full Bayesian implementation of the new npLFM family. Finally, applications of time-dependent npLFMs are shown via simulated and real data analysis. 

In Section~\ref{sec:bg}, finite and nonparametric LFMs are described. Section~\ref{sec:ibp} discusses the Indian Buffet Processes that form the kernel for the new class of npLFMs introduced in Section~\ref{sec:model_long}. Section~\ref{sec:inf} details the inference methods used to implement the models in Section~\ref{sec:model_long}, followed by synthetic and real data illustrations in Section~\ref{sec:results}. A brief discussion in Section~\ref{sec:conc} concludes the paper.

\section{Latent Factor Models}\label{sec:bg}
A Latent Variable Model (LVM) posits that the variation within a dataset of size $N$ could be described using some set of $K$ features, with each observation associated with a single parameter. As an example, consider a mixture of $K$ Gaussian distributions where each datum belongs to one of the $K$ mixture components parametrized by different means and variances. These parameters, along with the cluster allocations, comprise the latent variables. In alternative settings, the number of features may be infinite; however since each data point is associated with a single feature, the number of features required to describe the dataset will always be upper bounded by $N$.

While mixture models are widely used for representing the latent structure of a dataset, there are many practical applications where the observed data exhibit multiple underlying features. For example, in image modeling we may have two pictures, one of a dog beside a tree, and one of a dog beside a bicycle. If we assign both images to a single cluster, we ignore the difference between tree and bicycle. If we assign them to different clusters, we ignore the commonality of the dogs. In these situations, LVMs should allow each datum to be associated with multiple latent variables.

If each datum can be subdivided into a collection of discrete observations, one approach is to use an admixture model, such as latent Dirichlet allocation \citep{Blei:Ng:Jordan:2003} or a hierarchical Dirichlet process \citep{Teh:Jordan:Beal:Blei:2006}. Such approaches model the constituent observations of a data point using a mixture model, allowing a data point to express multiple features. For example, if a datum is a text document, the constituent observations might be words, each of which can be associated with a separate latent variable.

If it is not natural to split a data point into constituent parts---for example, if a data point is described in terms of a single vector---then we can construct models that directly associate each data point with multiple latent variables. This extension of LVMs is typically referred to as Latent Feature Models or Latent Factor Models (LFMs). For clarity, throughout this paper, LVM refers exclusively to models where each datum is associated with a single latent parameter, and LFM refers to models where each datum is associated with multiple latent parameters.

A classic example of an LFM is Factor Analysis \citep{Cattell:1952}, wherein one assumes $K$ $D$-dimensional latent features (or factors) $f_k$ which are typically represented as a $K\times D$ matrix $F$. Each datum, $x_n$, is associated with a vector of weights, $\lambda_n$, known as the factor loading, which determines the degree to which the datum exhibits each factor. Letting $X = (x_n)_{n=1}^N$ be the $N\times D$ data matrix and $L = (\lambda_n)_{n=1}^N$ be the $N\times K$ factor loading matrix, we can write $ X = L F + \mathbf{e}$, 
where $\mathbf{e}$ is a matrix of random noise terms. Factor Analysis can be cast in a Bayesian framework by placing appropriate priors on the factors, loadings and noise terms \citep{Press:Shigemasu:1989}. Such analysis is used in many contexts; as examples: micro array data \citep{Hochreiter:Clevert:Obermayer:2006}, dietary patterns \citep{Venkaiah:Brahmam:Vijayaraghavan:2011}, and psychological test responses \citep{Tait:1986}. Independent Component Analysis \citep[ICA]{Hyvarinen:Karhunen:Oja:2001} is a related model with independent non-Gaussian factors; ICA is commonly used in blind source separation of audio data.

A serious disadvantage of LFMs such as Factor Analysis and ICA is that they assume a fixed, finite number of latent factors. In many settings, such an assumption is hard to justify. Even with a fixed, finite dataset, picking an appropriate number of factors, \emph{a priori}, requires expensive cross-validation. In an online setting, where the dataset is constantly growing, it may be unreasonable to consider any finite upper bound. As illustrations, the number of topics that may appear in a newspaper, or the number of image features that may appear in an online image database, could grow unboundedly over time. One way of obviating this difficulty is to allow an infinite number of latent features \emph{a priori}, and to ensure that every datum exhibits only a finite number of features wherein  popular features tend to get reused. Such a construction would allow the number of exhibited features to grow in an unbounded manner as sample size grows, while still borrowing (statistical) strength from repeated features.

The transition from finite to infinite dimensional latent factors implies that the probability distributions on these factors in the generative process would now be elements in some function space; i.e., we enter the realm of Bayesian nonparametric inference. There is a vast literature on Bayesian nonparametric models; the classic references are \citet{Ferguson:1973} and \citet{Lo:1984}. Since the Indian buffet process is central to this paper, it is discussed in the following subsection.


\subsection{The Indian Buffet Process (IBP)}\label{sec:ibp}


A new class of nonparametric distributions of particular relevance to LFMs was developed by  \citet{Griffiths:Ghahramani:2005} who labeled their stochastic process prior  the Indian Buffet Process (IBP). This prior adopts a Bayesian nonparametric inference approach to the generative process of an LFM where the goal of unsupervised learning is to discover the latent variables responsible for generating the observed properties of a set of objects.

The IBP provides a mechanism for selecting overlapping sets of features. This mechanism can be broken down into two components: a  global random sequence of feature probabilities that assigns probabilities to infinitely many features, and a local random process that selects a finite subset of these features for each datum. 

The global sequence of feature probabilities is distributed according to a stochastic process known as the beta process \citep{Hjort:1990,Thibaux:Jordan:2007}. Loosely speaking, the beta process is a random measure, $B = \sum_{k=1}^\infty \mu_k \delta_{\theta_k}$, that assigns finite mass to a countably infinite number of locations; these atomic masses $\mu_k$ are independent, and are distributed according to the infinitesimal limit of a beta distribution. The locations, $\theta_k$, of the atoms parametrize an infinite sequence of latent features.

The subset selection mechanism is a stochastic process known as the Bernoulli process \citep{Thibaux:Jordan:2007}. This  process samples a random measure $\zeta_n = \sum_{k=1}^\infty z_{nk} \delta_{\theta_k}$, where each $z_{nk}\in \{0,1\}$ indicates the presence or absence of the $k$th feature $\theta_k$ in the latent representation of the $n$th datum, and are sampled independently as $z_{nk}\sim \mbox{Bernoulli}(\mu_k)$. We can use these random measures $\zeta_n$ to construct a binary feature allocation matrix $Z$ by ordering the features according to their popularity and aligning the corresponding ordered vector of indicators. This matrix will have a finite but unbounded number of columns with at least one non-zero entry; the re-ordering allows us to store the non-zero portion of the matrix in memory. It is often convenient to work directly with this random, binary matrix, and doing so offers certain insights into the properties of the IBP. This representation depicts the IBP as a (stochastic process) prior probability distribution over equivalence classes of binary matrices with a specified number of rows and a random, unbounded number of non-zero columns that grows in expectation as the amount of data increases. 

Consider a mathematical representation of the above discussion. Let $Z$ denote a random, binary matrix with $N$ rows and infinitely many columns, $K$ of which contain at least one non-zero entry. Then, following \citet{Griffiths:Ghahramani:2005}, the IBP prior distribution for $Z$ is given by
\begin{equation}
p(Z) = \frac{\alpha^K}{\prod_{h=1}^{2^N-1}K_h!}\exp\{-\alpha H_N\}\prod_{k=1}^K\frac{(N-m_k)!(m_k-1)!}{N!}
\label{eqn:ibpjoint}
\end{equation}
where $m_k = \sum_{n=1}^N z_{nk}$ is the number of times we have seen feature $k$; $K_h$ is the number of columns whose binary pattern encodes the number $h$ written as a binary number; $H_N$ is the $N$th harmonic number, and $\alpha>0$ is the parameter of the process. Succinctly, Equation~\ref{eqn:ibpjoint} is stated as: $Z \sim \mbox{IBP}(\alpha)$, where $\alpha$ is the parameter of the process; that is, Z has an IBP distribution with parameter $\alpha$.

What is the meaning of $\alpha$? Perhaps the most intuitive way to understand the answer to this question is to recast $p(Z)$ in Equation~\ref{eqn:ibpjoint} through the spectrum of an Indian buffet restaurant serving an infinite number of dishes at a buffet. Customers (observations) sequentially enter this restaurant, and select a subset of the dishes (observations). The first customer takes $\mbox{Poisson}(\alpha)$ dishes. The $n$th customer selects each previously sampled dish with probability $m_k/n$, where $m_k$ is the number of customers who have previously selected that dishes -- i.e. she chooses dishes proportional to their popularity. In addition, she samples a $\mbox{Poisson}(\alpha/n)$ number of previously untried dishes. This process continues until all $N$ customers visit the buffet. Now, represent the outcome of this buffet process in a binary matrix $Z$ where the rows of the matrix are customers and the columns are the dishes. The element $z_{n,k}$ is 1 if observation $n$ possesses feature $k$. Then, after some algebra, it follows that the probability distribution over the random, binary matrix $Z$ (up to a reordering of the columns) induced by this buffet process is invariant to the order in which customers arrived at the buffet, and is the expression given in Equation~\ref{eqn:ibpjoint}.

The meaning of $\alpha$ is now clear. The smaller the $\alpha$, the lower the number of features with $\sum_n z_{nk} > 0$, and the lower the average number of features per data point, with the number of features per datapoint distributed (marginally) as $\mbox{Poisson}(\alpha)$. Thus when the IBP is used in the generative process of an LFM, the total number of features exhibited by $N$ data points will be finite, but random, and this number will grow in expectation with the number of data points. This subset selection procedure behaves in a ``rich-get-richer'' manner--- if a dish had been selected by previous customers, it would likely be selected by new arrivals to the buffet. Stating generically, therefore, if a feature appears frequently in previously observed data points it would likely continue to appear again in subsequent observations as well.

We could use the IBP as the basis for an LFM by specifying a prior on the latent factors (henceforward denoted by a $ K \times D $ matrix $A$), as well as a likelihood model for generating observations, as shown in the following examples. If the data are real-valued vectors, an appropriate choice for the likelihood model data could be a weighted superposition of Gaussian features:
\begin{equation}
\begin{aligned}
Z = (Z_n)_{n=1}^N \sim& \mbox{IBP}(\alpha) & &\\
y_{nk} \sim& f &&\\
A_k \sim& \mbox{Normal}(0, \sigma_A^2I), &k&=1,2,\dots \\
X_n \sim& \mbox{Normal}((Z\circ Y)A, \sigma_X^2I), &n&=1,\dots, N.\label{eqn:lingauss}
\end{aligned}
\end{equation}

Here, $Y$ is the $N\times \infty$ matrix with elements $y_{nk}$; $A$ is the $\infty \times D$ matrix with rows $A_k$; $\circ$ is the Hadamard product; and  $f$ is a distribution over the weights for a given feature instance. Note that, while we are working with infinite-dimensional matrices, the number of non-zero columns of $Z$ is finite almost surely, so we only need to represent finitely many columns of $Y$ and rows of $A$. If $f=\delta_{1}$, we have a model where features are either included or not in a data point, and where a feature is the same each time it appears; this straightforward model was proposed by \citet{Griffiths:Ghahramani:2005}, but is somewhat inflexible for real-life modeling scenarios.

Letting $f=\mbox{Normal}(\mu_f, \sigma_f^2)$ gives Gaussian weights, yielding a nonparametric variant of Factor Analysis \citep{Knowles:Ghahramani:2007,Teh:Gorur:Ghahramani:2007}. This approach is useful in modeling psychometric test data, or analyzing marketing survey data. Letting $f=\mbox{Laplace}(\mu_f, b_f)$ results in a heavier-tailed distribution over feature weights, yielding a nonparametric version of Independent Components Analysis \citep{Knowles:Ghahramani:2007}. This allows one to perform blind source separation where the number of sources is unknown, making it a potentially useful tool in signal processing applications.

Often, one encounters binary-valued data: for example, an indicator vector corresponding to disease symptoms (where a 1 indicates the patient exhibits that symptom), or purchasing patterns (where a 1 indicates that a consumer has purchased that product). In these cases, a weighted superposition model is not directly applicable, but it may be reasonable to believe  there are multiple latent causes influencing whether an element is turned on or not. One option in such cases is to use the IBP with a likelihood model \citep{Wood:Griffiths:Ghahramani:2006} where observations are generated according to:
\begin{equation*}
\begin{split}
Z = (Z_n)_{n=1}^N \sim& \mbox{IBP}(\alpha)\\
y_{dk} \sim& \mbox{Bernoulli}(p)\\
P(x_{nd}=1|Z,Y) =& 1-(1-\lambda)^{Z_nY_d^T}(1-\epsilon),
\end{split}
\end{equation*}
and where $Y$ is the $D\times \infty$ matrix with elements $y_{dk}$; $Z_i$ and $Y_i$ are the $i$th rows of $Z$ and $Y$ respectively. 

The above illustrations exemplify the value of IBP priors in LFMs. While these illustrations cover a vast range of applied problems, there are limitations. Notable among them is that the above LFMs do not encapsulate time dynamics. The aim of this paper is to develop a new family of IBP-based LFMs that obviates this crucial shortcoming. Additionally, unlike the afore-described models, the new class also allows one to capture repeat occurrence of a feature through time; i.e., \emph{persistence} of latent factors. (Recall from the Introduction the example of a flautist's musical note persisting in successive time intervals.)

\subsection{Temporal Dynamics in npLFMs}\label{sec:dynamic_bg}

The IBP, like its finite-dimensional analogues, assumes that the data are exchangeable. In practice, this could be a restrictive assumption. In many applications, the data exhibit either longitudinal (time) dependence or epidemiological dependence.  Since the latter form of dependency is not the focus of this paper, we no longer consider it in the ensuing discussions. Some important references for this latter type of dependency include \citet{Ren:Wang:Carin:Dunson:2011}, \citet{Foti:Futoma:Rockmore:Williamson:2013}, and  \citet{Zhou:Yang:Sapiro:Dunson:Carin:2011}.

Longitudinal dependence considers the case where each datum corresponds to an instantiation of a single evolving entity at different points in time. For example, data might correspond to timepoints in an audio recording, or measurements from a single patient over time. Mathematically, this means we would like to capture continuity of latent features. This setting has been considered less frequently in the literature. The Dependent Indian Buffet Process  \citep[DIBP,][]{Williamson:Orbanz:Ghahramani:2010} captures longitudinal dependence by modeling the occurrence of a given feature with a transformed Gaussian process. This allows for a great deal of flexibility in the form of dependence but comes at high computational cost: inference in each Gaussian process scales cubically with the number of time steps, and we must use a separate Gaussian process for each feature. 

Another model for longitudinal dependence is the Distance-Dependent Indian Buffet Process (DDIBP) of \citet{Gershman:Frazier:Blei:2015}. In this model, features are allocated using a variant of the IBP metaphor, wherein each pair of data points is associated with some distance measure. The probability of two data points sharing a feature depends on the distance between them. With an appropriate choice of distance measure, this model could prove useful for time-dependent data.

An alternative approach is provided by  IBP-based hidden Markov models. For example, the Markov IBP extends the IBP such that rows are indexed by time and the presence or absence of a feature at time $t$ depends only on which features were present at time $t-1$. This model is extended further in the Infinite Factorial Unbounded State Hidden Markov Model \citep[IFUHMM,][]{Valera:Ruiz:Perez:2016} and the Infinite Factorial Dynamical Model  \citep[IFDM,][]{Valera:Ruiz:Lennart:PerezCruz2015}. These related models combine hidden Markov models, one which controls which features are present, and one which controls the expression of that feature. The feature presence/absence is modeled using a Markov IBP \citep{VanGael:Teh:Ghahramani:2009}. At different time points, a single feature can have multiple expressions. During a contiguous time period where feature $k$ is present, it moves between these expressions using Markovian dynamics.  While this increases the model flexibility, this comes at a cost of interpretability. Unlike the DIBP, the DDIBP, and the model proposed in this paper, the IFUHMM and IFDM do not impose any similarity requirements on the expressions of a given feature and can therefore use a single feature to capture two very different effects, provided they never occur simultaneously.

While not a dynamic latent factor model, another dynamic model based on npLFMs is the beta process autoregressive hidden Markov model \citep[BP-AR-HMM,][]{Fox:Sudderth:Jordan:Willsky:2009,Fox:Hughes:Sudderth:Jordan:2014}. In this model, an IBP is used to couple multiple time-series in a vector autoregressive model. The IBP is used to control the weights assigned to the lagged components; these weights are stationary over time.

In addition to the longitudinally dependent variants npLFMs mentioned here, there also exist a large number of temporally dependent npLVMs. In particular, dependent Dirichlet processes \citep[e.g.][]{Maceachern:2000,Caron:Davy:Doucet:2017,Lin:Grimson:Fisher:2010,Griffin:2011} extend the Dirichlet process to allow temporal dynamics, allowing for time-dependent clustering models. Hidden Markov models based on the hierarchical Dirichlet process  \citep{Fox:Sudderth:Jordan:Willsky:2008,Fox:Sudderth:Jordan:Willsky:2011,Zhang:Guletkin:Paislet:2016} allow the latent variable associated with an observation to evolve in a Markovian manner. We do not discuss these methods in depth here, since they assume a single latent variable at each time point.

The model we propose in Section~\ref{sec:model_long} falls into this class of longitudinally dependent LFMs. Unlike the DIBP, DDIBP, our model explicitly models feature persistence. Unlike all the models described above, our model allows multiple instances of a feature to appear at once. This is appropriate in many contexts; for instance, in music analysis, where each note has an explicit duration and two musicians could play the same note simultaneously. Importantly, the proposed nonparametric LFM leaves the underlying IBP  mechanism intact, leading to more straightforward inference procedures when compared to DIBP and DDIBP.


\section{A New Family of npLFMs for Time-Dependent Data}\label{sec:model_long}

Existing temporally dependent versions of the IBP ~\citep{Williamson:Orbanz:Ghahramani:2010,Gershman:Frazier:Blei:2015} rely on explicitly or implicitly varying the underlying latent feature probabilities---a difficult task--- and inference tends to be computationally complex. 

Our proposed method obviates these  limitations. In a nutshell, unlike existing dependent npLFMs, we build our model on top of a single IBP, as described in Section~\ref{sec:ibp}. Temporal dependence is encapsulated via a \textit{likelihood model}. The value of our approach could be best understood via some simple examples. Consider audio data. A common approach to modeling audio data is to view them as superpositions of multiple sources; for example, individual speakers or different instruments. The IBP has previously been used in these types of applications  \citep{Knowles:Ghahramani:2007,DoshiVelez:2009}. However, these approaches ignore \emph{temporal dynamics} present in most audio data. Recall the flautist example: at very short time intervals, if a flautist is playing a B$\flat$ at time $t$, it is likely that note could still be playing at time $t+k$, $k=1,2,\dots$. Our proposed model captures this dependency in the musical notes. In Section~\ref{sec:results}, using real data, we show the benefit of incorporating this dynamic, temporal \emph{feature persistence} and contrast it to a static IBP, DIBP, and DDIBP. 

As noted in the Abstract, another illustration is the modeling of sensor outputs over time. Sensors record responses to a variety of external events: for example, in a building we may have sensors recording temperature, humidity, power consumption and noise levels. These are all altered by events happening in the building---the presence of individuals; the turning on and off of electrical  devices; and so on. Latent factors  influencing the sensor output are typically present for a contiguous period of time before  disappearing; besides, multiple factors could be present at a time. Thus, for instance, our model should capture the effect on power consumption due to an air conditioning unit being turned on from 9am to 5pm, and which could be subject to latent disturbances during that time interval such as voltage fluctuations. 

Consider a third illustration involving medical signals such as EEG or ECG data. Here, we could identify latent factors causing unexpected patterns in the data, as well as infer the duration of their influence. As in previous examples, we expect such factors to contribute for a contiguous period of time: for instance, a release of stress hormones would affect all time periods until the levels decrease below a threshold. Note that the temporal variation in all three illustrations above cannot be accurately captured with epidemiologically dependent factor models where the probability of a factor varies smoothly over time, but the actual presence or absence of that feature is sampled independently given appropriate probabilities. This approach would lead to noisy data where a feature randomly oscillates between on and off.

Under the linear Gaussian likelihood model described in Equation~\ref{eqn:lingauss}, conditioned on the latent factors $A_k$, the $n$th datum is characterized entirely by the $n$th row of the IBP-distributed matrix $Z$ thereby ensuring that the data, like the rows of $Z$, are exchangeable. In the following, the key point of departure from the npLFMs described earlier is this: we now let the $n$th datum depend not only on the $n$th row of $Z$, but also on the $n-1$ preceding rows, thus breaking the exchangeability of the $X_n$ data sequence. This is the mathematical equivalent of dependency in the data that we now formalize.

Associate each non-zero element $z_{nk}$ of $Z$ with a geometrically-distributed ``lifetime'', namely $\ell_{nk} \sim \mbox{Geometric}(\rho_k)$. An instance of the $k$th latent factor is then incorporated from the $n$th to the $(n+\ell_{nk}-1)$th datum. The $n$th datum is therefore associated with a set $\mathcal{Y}_n$, which represents the time series data, of feature indices $\{(i,j): z_{ij}=1, i+\ell>n\}$. We use the term ``feature'' to refer to a factor, and the term ``feature instance'' to refer to a specific realization of that factor. For example, if each factor corresponds to a single note in an audio recording, the global representation of the note $C$ would be a feature, and the specific instance of note $C$ that starts at time $n$ and lasts for a geometrically distributed time would be a feature instance. If we assume a shared lifetime parameter, $\rho_k=\rho$ for all features, then the number of features at any time point is given, in expectation, by a geometric series $E[|\mathcal{Y}_n|] = \sum_{i=0}^{n-1}\alpha \rho^i\rightarrow \frac{\alpha}{1-\rho}$ as $n\rightarrow \infty$, i.e. as we forget the start of the process. More generally, we allow $\rho_k$ to differ between features, and place a $\mbox{Beta}(a_\rho, b_\rho)$ prior on each $\rho_k$. By a judicious choice of the hyper-parameters, this prior could be easily tailored to encapsulate vague prior knowledge or contextual knowledge. (As an added bonus, it leads to simpler stochastic simulation methods which will be discussed later on.)

This geometric lifetime is the source of dependency in our new class of IBP-based npLFMs. It captures the idea of feature \textit{persistence}: a feature instance ``turned on'' at time $t$ appears in a geometrically distributed number of future time steps. Since any feature instance that contributes to $x_n$ also contributes to $x_{n+1}$ with probability $\rho_k$, we expect $x_n$ to share $\frac{\alpha+\rho-1}{1-\rho}$ feature instances with $x_{n-1}$, and to introduce $\alpha$ new feature instances. Of these new feature instances, we expect $\alpha/n$ to be versions of previously unseen features.

Note that this construction allows a specific datum to exhibit multiple instances of a given latent factor. For example, if $\mathcal{Y}_n=\{(n,1), (n,3),(n-1,1)\}$, then the $n$th datum will exhibit two copies of the first feature and one copy of the third feature. In many settings, this is a reasonable assumption: two trees appearing in a movie frame, or two instruments playing the same note at the same time.

The construction of dependency detailed above could now be combined with a variety of likelihood functions (or models) appropriate for different data sources or applications. We could also replace the geometric lifetime with other choices, for example using semi-Markovian models as in \citet{johnson2013bayesian}. Armed with this kernel of geometric dependency and likelihood functions, we now illustrate the broad scope of the proposed family of time-dependent npLFMs via two generalizations. Later, we demonstrate these using real or synthetic data.

Adapting the linear Gaussian IBP LFM used by \citet{Griffiths:Ghahramani:2005} to our dynamic time-dependent model, where each datum is given by a linear superposition of Gaussian features, results in:
\begin{equation}
\begin{aligned}
Z \sim& \mbox{IBP}(\alpha) && \\
A_k \sim& \mbox{Normal}(0,\sigma_A^2) && \\
\ell_{nk}\sim& \mbox{Geometric}(\rho_k), &k&=1,2,\dots\\
\mu_n =& \textstyle\sum_{i=1}^n \sum_{k=1}^\infty z_{ik}I(i+\ell_{ik}>n)A_k & &\\
X_n \sim& \mbox{Normal}(\mu_k, \sigma_X^2), &n&=1,\dots, N,\label{eqn:dynamic_general}
\end{aligned}
\end{equation}
where $I()$ is the indicator function.

Consider a second generalization where one wishes to model variations in the appearance of a feature. Here, we can customize each feature instance using a feature weight $b_{nk}$ distributed according to some distribution $f$ so that: 
\begin{equation}
\begin{aligned}
Z \sim& \mbox{IBP}(\alpha) &&\\
A_k \sim& \mbox{Normal}(0,\sigma_A^2) &&\\
\ell_{nk}\sim& \mbox{Geometric}(\rho_k), &k&=1,2,\dots\\
b_{nk}\sim& f &&\\
\mu_n =& \textstyle \sum_{i=1}^n \sum_{k=1}^\infty z_{ik}b_{ik}I(i+\ell_{ik}>n)A_k&&\\
X_n \sim& \mbox{Normal}(\mu_k, \sigma_X^2), &n&=1,\dots, N.
\end{aligned}
\label{eqn:dynamic_amplitude}
\end{equation}

For example, in modeling audio data, a note or chord might be played at different volumes throughout a piece. In this case, it is appropriate to incorporate a per-factor-instance gamma-distributed weight, $b_{nk}\sim \mbox{Gamma}(\alpha_B,\beta_B)$. 

The new family of time-dependent models above could be used in many applications, provided they are computationally feasible. In the following, we develop stochastic simulation methods to achieve this goal.

\section{Inference Methods for npLFMs}\label{sec:inf}
A number of inference methods have been proposed for the IBP, including Gibbs samplers \citep{Griffiths:Ghahramani:2005,Teh:Gorur:Ghahramani:2007}, variational inference algorithms \citep{Doshi:Miller:VanGael:Teh:2009}, and sequential Monte Carlo samplers \citep{Wood:Griffiths:2006}. In this work, we focus on Markov chain Monte Carlo (MCMC) approaches (like the Gibbs sampler) since, under certain conditions, they are guaranteed to asymptotically converge to the true posterior distributions of the random parameters. Additionally, having tested various simulation methods for the dynamic models introduced in this paper, we found that the MCMC approach is easier to implement, and has good mixing properties.

When working with nonparametric models, we are faced with a choice. One, perform inference on the full nonparametric model by assuming infinitely many features {\emph{a priori} and inferring the appropriate number of features required to model the data.   Two, work with a large, $K$-dimensional model that converges (in a weak-limit sense) to the true posterior distributions as $K$ tends to infinity. The former approach will asymptotically sample from the true posterior distributions, but the latter approximation approach is often preferred in practice due to lower computational costs. We describe algorithms for both approaches.

	\subsection{An MCMC Algorithm for the Dynamic npLFM}\label{sec:MCMCbasic}
	Consider the weighted model in Equation~\ref{eqn:dynamic_amplitude}, where the feature instance weights $b_{nk}$ are distributed according to some arbitrary distribution $f(b)$ defined on the positive reals. Define $B$ as the matrix with elements $b_{nk}$.  Inference for the uniform-weighted model in Equation~\ref{eqn:dynamic_general} is easily recovered by setting $b_{nk}=1$ for all $n,k$.
	
	Our algorithms adapt existing fully nonparametric \citep{Griffiths:Ghahramani:2005,DoshiVelez:Ghahramani:2009} and weak-limit MCMC algorithms \citep{zhou2009non} for the IBP. One key difference is that we must sample not only whether feature $k$ is instantiated in observation $n$, but also for the number of observations for which this particular feature remains active. We obtain inferences for the IBP-distributed matrix $Z$ and the lifetimes $\ell_{nk}$ using a Metropolis-Hastings algorithm described below.

	
	\paragraph{Sampling $Z$ and the $\ell_{nk}$ in the Full Nonparametric Model:}
	
	We jointly sample the feature instance matrix $Z$ and the corresponding lifetimes $\ell$ using a slice sampler \citep{Neal:2003}. Let $\Lambda$ be the matrix whose elements are given by $\lambda_{nk}:=z_{nk}\ell_{nk}$. To sample a new value for  $\lambda_{nk}$ where $\sum_{i\neq n}\lambda_{ik}>0$, we first sample an auxiliary slice variable $u\sim \mbox{Uniform}(0,Q^*(\lambda_{nk}))$, where $Q^*(\lambda_{nk}) = p(\lambda_{nk}|\rho_k, m_k^{-n})p(X|\lambda_{nk}, A,B, \sigma_X^2)$. Here, the likelihood term $p(X|\lambda_{nk}, A, B, \sigma_X^2)$ depends on the choice of likelihood, and
	
	\begin{equation}
	p(\lambda|\rho_k m_k^{-n}) = \begin{cases} \frac{N-m_k^{-n}}{N} & \mbox{if }\lambda=0\\
	\frac{m_k^{-n}}{N} \rho(1-\rho)^{\lambda_{nk}-1} & \mbox{otherwise}
	\end{cases}\label{eqn:plambda_np}
	\end{equation}
	
	We then define a bracket centered on the current value of $\lambda_{nk}$, and sample $\lambda_{nk}^*$ uniformly from this bracket. We accept $\lambda_{nk}^*$ if $Q(\lambda_{nk}^*) = p(\lambda_{nk}^*|\rho_k, m_k^{-n})p(X|\lambda_{nk}^*, A,B, \sigma_X^2) > u$. If we do not accept $\lambda^*_{nk}$, we shrink our bracket so that it excludes $\lambda_{nk}^*$ but includes $\lambda_{nk}$, and repeat this procedure until we either accept a new value, or our bracket contains only the previous value.
	
	For the $n$th row of $Z$, we can sample the number of singleton features --- i.e. features where $z_{nk}=1$ but $\sum_{i\neq n}z_{ik}=0$ --- using a Metropolis Hastings step. We sample the number $K^*$ of singletons in our proposal from a $\mbox{Poisson}(\alpha/N)$ distribution, and sample corresponding values of $b^*_{nk}\sim f(b)$. We also sample corresponding lifetime probabilities $\rho_k^*\sim \mbox{Beta}(a_\rho, b_\rho)$ and lifetimes $\ell_{nk}^* \sim \mbox{Geometric}(\rho_k^*)$ for the proposed singleton features. We then accept the new $\Lambda$ and $B$ with probability
	$$\min\left(1, \frac{Q(X|\Lambda^*, A,B^*, \sigma_X)}{Q(\Lambda, A, B, \sigma_X)}\right),$$
	for some proposal distribution $ Q $.
	
	\paragraph{Sampling $Z$ and the $\ell_{nk}$ using a Weak-Limit Approximation:}
	Inference in the weak-limit setting is more straightforward since we do not have to worry about adding and deleting new features. We modify the slice sampler for the full nonparametric model, replacing the definition of $p(\lambda_{nk}|\rho_k, m_{k}^{-n})$ in Equation~\ref{eqn:plambda_np} by
	\begin{equation}
	p(\lambda|\rho_k m_k^{-n}) = \begin{cases} \frac{N-m_k^{-n}}{N+\frac{\alpha}{K}} & \mbox{if }\lambda=0\\
	\frac{m_k^{-n}+\frac{\alpha}{K}}{N+\frac{\alpha}{K}} \rho(1-\rho)^{\lambda_{nk}-1} & \mbox{otherwise,}
	\end{cases}\label{eqn:plambda_wl}
	\end{equation}
	and by slice sampling $\lambda_{nk}$ even if $\sum_{i\neq n}\lambda_{ik}=0$. In the weak limit setting, we do not have a separate procedure for sampling singleton features.

	\paragraph{Sampling $A$ and $B$:}
	Conditioned on $Z$ and the $\ell_{nk}$, inferring $A$ and $B$ will generally be identical to a model based on the static IBP, and does not depend on whether we used a weak-limit approximation for sampling $Z$. Recall that $\mathcal{Y}_{n}$ is the vector of feature indices $\{(i,j): z_{ij}=1, i+\ell>n\}$. Let $Y$ be the matrix with elements $y_{nk} = \sum_{i: (i,k)\in \mathcal{Y}_{n}} b_{nk}$ -- i.e. the total weight given to the $k$th feature in the $n$th observation. Then conditioned on $Y$ and $B$, the feature matrix $A$ is normally distributed with mean
	$$\mu_A = \left(Y^TY + \frac{\sigma_X^2}{\sigma_A^2}\mathbf{I}\right)^{-1}Y^TX$$
	and block-diagonal covariance, with each column of $A$ having the same covariance 
	$$\Sigma_A = \sigma_x^2\left(Y^T Y+\frac{\sigma_X^2}{\sigma_A^2}\mathbf{I}\right)^{-1}.$$
	
	We can use a Metropolis-Hastings proposal to sample from the conditional distribution  $P(b_{nk}|X,Z,\{\ell_{nk}\}, A, \sigma_X^2)$ --- for example, sampling $b_{nk}^*\sim f(b)$ and accepting with probability
	$$\min\left(1, \frac{P(X|Z, \{\ell_{nk}\}, A, B^*, \sigma_X)}{P(X|Z,\{\ell_{nk}\}, A,B, \sigma_X)}\right).$$
	
	\paragraph{Sampling Hyperparameters:}
	With respect to the choice of model we could either incorporate informative prior beliefs or use non-informative settings, depending on the user knowledge and the data at hand. Without loss of generality, we place inverse gamma priors on $\sigma_X^2$ and $\sigma_A^2$ and beta priors on each of the $\ell_k$; then, we can easily sample from their conditional distributions due to conjugacy. 
	Similarly, if we place a $\mbox{Gamma}(a_\alpha,b_\alpha)$ prior on $\alpha$, we can sample from its conditional distribution
	$$\alpha|Z\sim \mbox{Gamma}\left(K+a_\alpha, \frac{b_{\alpha}}{1+b_\alpha H_n}\right)$$
	where $H_n$ is the $n$th harmonic number. These inverse gamma and gamma  prior distributions are general since, by a judicious choice of hyperparameter values, they could be tailored to model very little to strong prior information. 
	
	\section{Experimental Evaluation}\label{sec:results}
	Here the proposed models and stochastic simulation methods are exemplified via synthetic and real data illustrations. In the synthetic illustration, we used the full nonparametric simulation method; in the real data examples, we used the weak-limit approximation version of the MCMC algorithm to sample the nonparametric component. We do this to allow fair comparison with the DIBP and DDIBP, which both use a weak-limit approximation. We choose to compare with the IFDM over the related IFUHMM since it offers a more efficient inference algorithm, and because code was made available by the authors.
	
	The ``gold standard'' in assessing npLFMs is to first set aside a hold-out sample. Then, using the estimated parameters one predicts these held-out data; i.e., comparing actual versus predicted values. In this section, we do this by alternately imputing the missing values from their appropriate conditional distributions, and using the imputed values to sample the latent variables.

	Since the aim is to compare static npLFM models and existing dynamic (DIBP, DDIBP and IFDM) models with the temporal dynamic npLFM models developed in this paper, the mean square error (MSE) is used to contrast the performance of these approaches on the held-out samples. We choose to consider squared error over absolute error due to its emphasis on extreme values. In the interest of space, we have not included plots or figures demonstrating the mixing of the MCMC sampler though one may use typical MCMC convergence heuristics to assess convergence \citep[for example]{Geweke:1991,Gelman:Rubin:1992,Brooks:Gelman:1998}.
	
	
	\subsection{Synthetic Data}
	
	To show the benefits of explicitly addressing temporal dependence, we carried out the following.
	
	\begin{itemize}
		\item Generate a synthetic dataset with the canonical ``Cambridge bars'' features shown in Figure~\ref{fig:synthA}; these features were used to generate a longitudinally varying dataset. 
		\item Simulate a sequence of $N=500$ data points corresponding to $N$ time steps.
		\item  For each time step, add a new instance of each feature with probability 0.2, then sample an active lifetime for that feature instance according to a geometric distribution with parameter 0.5.
		\item Each datum was generated by superimposing all the active feature instances (i.e. those whose lifetimes had not expired) and adding Gaussian noise to give an $6\times 6$ real-valued image.
		\item We designated 10\% of the observations as our test set. For each test set observation, we held out 30 of the 36 pixels. The remaining 6 pixels allowed us to infer the features associated with the test set observations.
	\end{itemize}

	\begin{figure}
		\centering
		\includegraphics[width=.45\textwidth]{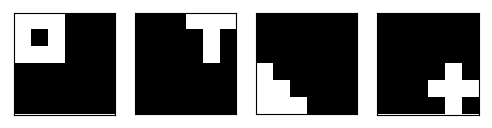}\\
		\includegraphics[width=1.0\textwidth]{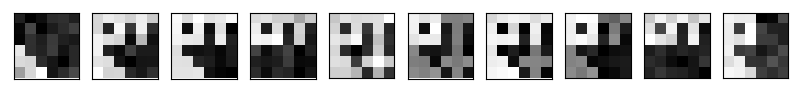}
		\caption{Top row: Four synthetic features used to generate data. Bottom row: Ten consecutive observations.}
		\label{fig:synthA}
	\end{figure}
	
	We considered four models: The static IBP; the dynamic npLFM proposed in this paper; the DIBP, the DDIBP, and the IFDM. For the dynamic npLFM and the static IBP, we used our fully nonparametric sampler. For the DIBP, DDIBP and the IFDM we used code available from the authors. The DIBP and DDIBP codes use a weak limit sampler; we fixed $K=20$ for the DDIBP and for the DIBP; the lower value for the DIBP is needed due to the much slower computational protocol for this method. 
	
	Table~\ref{table:synthetic1} shows the MSEs obtained on the held-out data; the number of features; and the average feature persistence. All values are the final MSE averaged over 5 trials from the appropriate posterior distributions following convergence of the MCMC chain. The average MSE is significantly lower for our dynamic model in contrast to all the other models we compared against. 
	Next, consider Figure~\ref{fig:instances} that shows the total number of times each feature contributes to a data point (i.e., the sum of that feature's lifetimes), based on a single iteration from both the dynamic and the static model. It is clear that the dynamic model reuses common features a larger number of times than the static model.
	\begin{table}
		\centering
		\begin{tabular}{|c|c|c|c|}
			\hline
			& MSE & Number of features & Average persistence \\
			\hline
			Dynamic npLFM & $0.274\pm 0.02$ & $15.80 \pm 0.748$ & $2.147 \pm 0.58$\\
			\hline
			Static npLFM & $0.496 \pm 0.04$ & $19.80 \pm 0.400$ & $-$ \\
			\hline
			DIBP & $0.459 \pm 0.01$ & $20^*$ & $-$ \\
			\hline
			DDIBP & $0.561 \pm 0.02$ & $20^{\dagger}$ & $-$\\
			\hline
			IFDM & $0.7513 \pm  0.003$ & $2^{\dagger}$ & $-$ \\
			\hline 
		\end{tabular}
		\caption{Average MSE; number of features; and feature persistence on synthetic data under static and dynamic npLFMs. $ ^*$Note that the DIBP was restricted to $K=20$ features for computational reasons. $^{\dagger}$The results over the 5 trials resulted in learning the same number of features.}
		\label{table:synthetic1}
	\end{table}
	
	There are two critical reasons for this superior performance. First, consider a datum with two instances of a given feature: one that has just been introduced, and one that has persisted from a previous time-point. Our dynamic model is able to use the same latent feature to model both feature instances, while the static model, the DIBP, and the DDIBP must use two separate features (or model this double-instance as a separate feature from a single-instance). This is seen in the lower average number of features required by the dynamic model (Table~\ref{table:synthetic1}), and in the greater number of times common features are reused (Figure~\ref{fig:instances}).
	
	In general, if (in the limit of infinitely many observations) there is truely a finite number of latent features, it is known that non-parametric models will tend to overestimate this number \citep{Miller:Harrison:2013}. With that said, from a modeling perspective we generally wish to recover fewer redundant features, giving a parsimonious reconstruction of the data. We can see that we achieve this, by comparing the number and populatity of the features recovered with our dynamic model, relative to the static model.
	
	Second, the dynamic npLFM makes use of the ordering information and anticipates that feature instances will persist for multiple time periods; this means that the latent structure for a given test-set observation is informed not only by the six observed pixels, but also by the latent structures of the adjacent observations.  We see that the average feature persistence is $2.147$ time steps,  which confirms that the dynamic model makes use of the temporal dynamics inherent in the data. While the DIBP, DDIBP and IFDM both have mechanisms to model temporal variation, their models do not match the method used to generate the data, and cannot capture the persistence variation explicitly.
	\begin{figure}
		\centering
		\includegraphics[width=.65\textwidth]{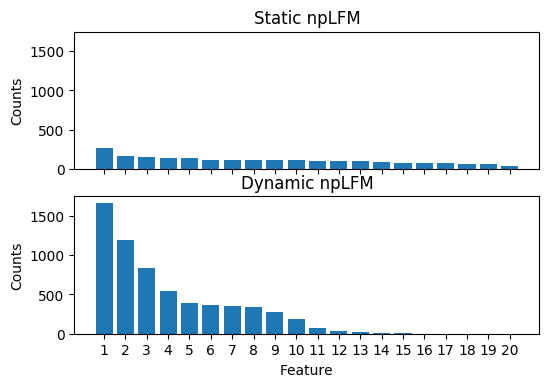}
		\caption{Number of times each feature contributes to a data point under static and dynamic npLFMs. Note that under the dynamic model, a feature can contribute multiple times to the same data point. In this setting, the features are arbitrarily labeled and thus labeled according to their popularity.}
		\label{fig:instances}
	\end{figure}
	
	\begin{table}
		\centering
		\begin{tabular}{|c|c|c|c|c|}
			\hline
			& Household power consumption & Audio data & Bird call data \\
			\hline
			Dynamic npLFM & $0.287 \pm 0.013$ & $0.722 \pm 0.007$ & $ 0.561 \pm 0.306$\\
			\hline
			Static npLFM & $1.835 \pm 0.182$ &  $1.013 \pm 0.013$ & $ 1.026 \pm 0.481 $ \\ 
			\hline
			DDIBP & $1.424 \pm 0.069$ & $1.289 \pm 0.224$ & $ 0.606 \pm 0.036 $ \\ 
			\hline
			DIBP & $1.324 \pm 0.106$ & $ 1.845 \pm 0.264$ & $ 1.308 \pm 0.744 $ \\
			\hline
			IFDM & $ 0.294 \pm  0.022$ & $ 1.906 \pm 0.009 $ & $1.222 \pm 0.130$ \\
			\hline
		\end{tabular}
		\caption{Average MSE obtained on the empirical datasets by the dynamic model proposed in this paper; the static IBP latent feature model; the DDIBP; the DIBP; and the IFDM.}
		\label{table:real1}
	\end{table}

	\subsection{Household Power Consumption Real Data Illustration}
	A number of different appliances contribute to a household's overall power consumption, and each appliance will have different energy consumption and operating patterns. We analyzed the ``Individual household electric power consumption'' data set\footnote{We only analyzed a subset of the data for computational reasons.} available from the UCI Machine Learning Repository\footnote{\texttt{http://archive.ics.uci.edu/ml}}. This dataset records overall minute-averaged active power, overall minute-averaged reactive power,  minute-averaged voltage, overall minute-averaged current intensity, and watt-hours of active energy on three sub-circuits within one house.
	
	We examined 500 consecutive recordings. For each recording, we independently scaled each observed feature to have zero mean and unit variance, and subtracted the minimum value for each observed feature. The preprocessed data can, therefore, be seen as excess observation above a baseline, with all features existing on the same scale justifying a shared prior variance. Based on the assumption that a given appliance's energy demands are approximately constant, we applied our dynamic npLFM with constant weights,  described in Equation~\ref{eqn:dynamic_general}.
	\begin{figure}[h]
		\centering
		\includegraphics[width=\textwidth]{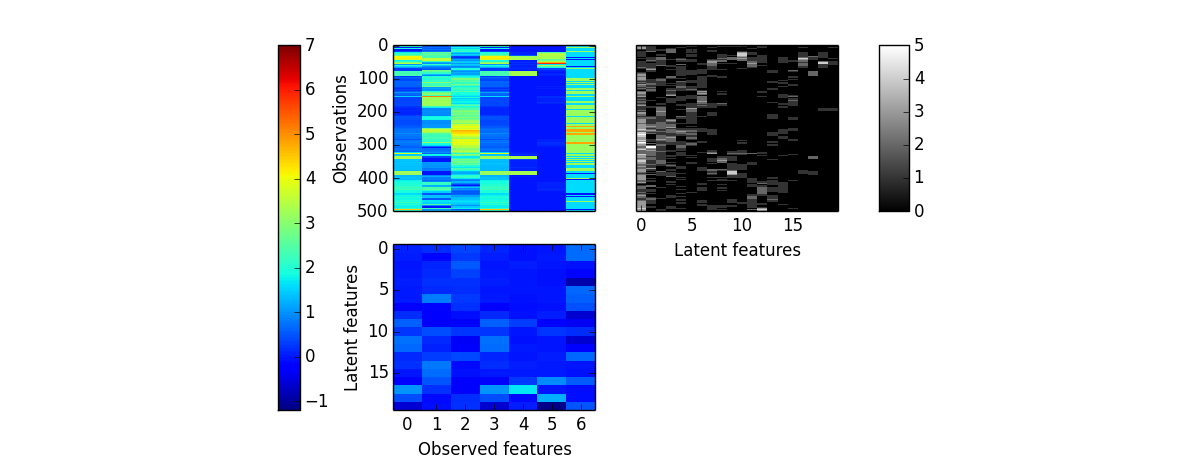}
		\caption{Latent structure obtained from the household power consumption data using the dynamic npLFM. Top left: Intensity of observed feature at each observation (after pre-processing). Bottom left: Latent features found by the model. Top right: Number of instances of each latent feature, at each observation.}
		\label{fig:hpc_dynamic}
	\end{figure}

	\begin{figure}[h]
		\centering
		\includegraphics[width=\textwidth]{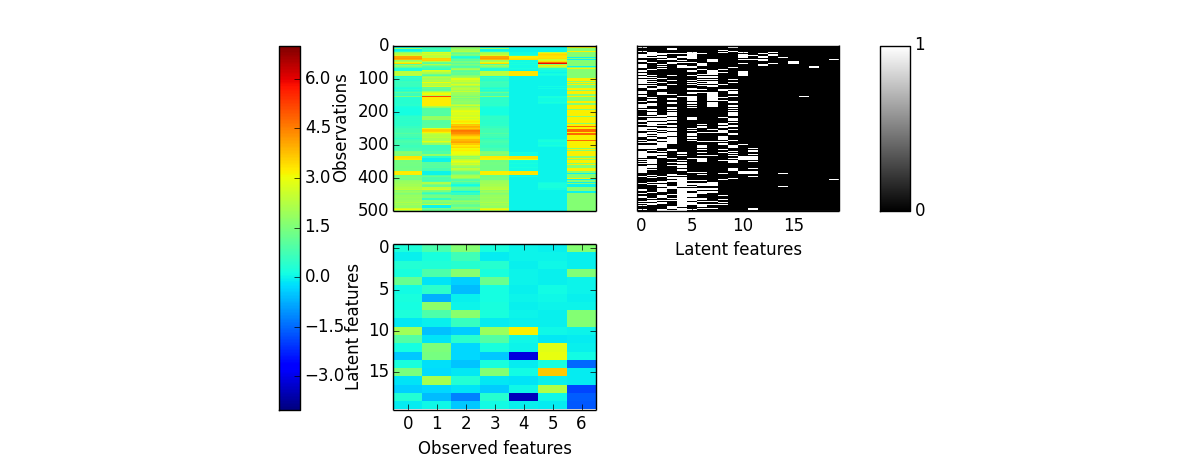}
		\caption{Latent structure obtained from the household power consumption data using the static IBP. Top left: Intensity of observed feature at each observation (after pre-processing). Bottom left: Latent features found by the model. Top right: Number of instances of each latent feature, at each observation.}
		\label{fig:hpc_ibp}
	\end{figure}
	
	\begin{figure}[h]
		\centering
		\includegraphics[width=.8\textwidth]{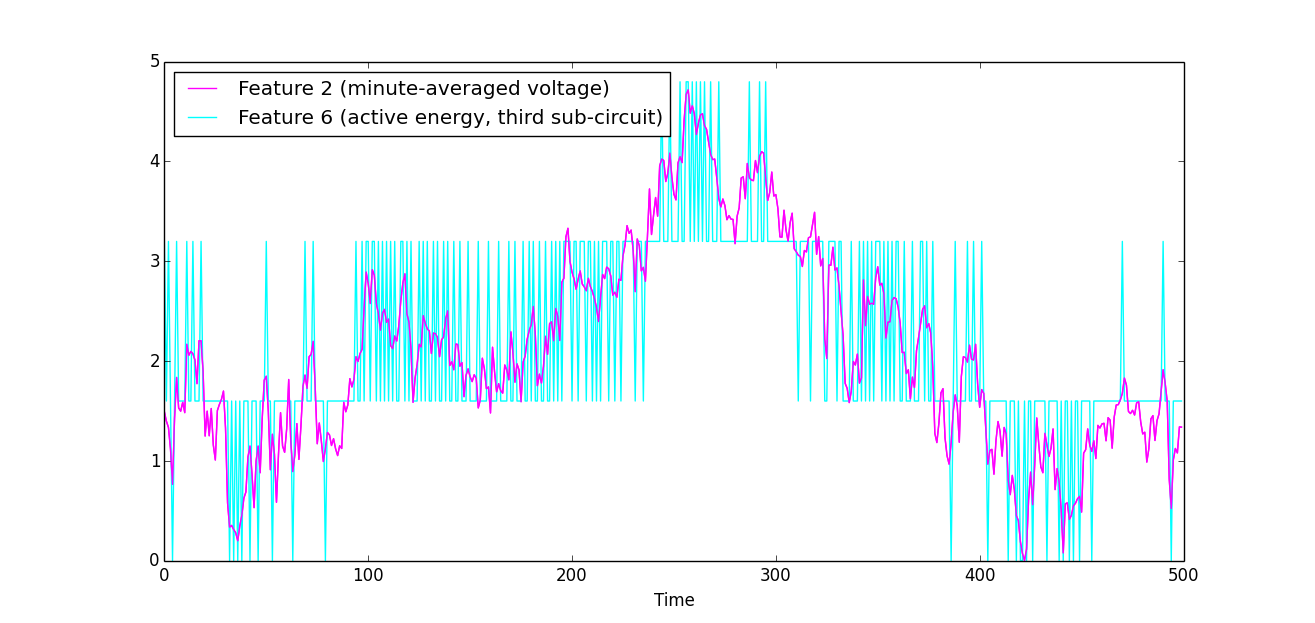}
		\caption{Plot of observed features 2 and 6 from the household power consumption data, over time.}
		\label{fig:time_series}
	\end{figure}
	We again compared against a static IBP, the DIBP, the DDIBP (with exponential similarity measure), and the IFDM. For all models, we used a weak limit sampler with a maximum of 20 features. For validation, 10\% of the data were set aside, with a randomly selected six out of seven dimensions being held out. We expect the dynamic models to perform better than the static model, given the underlying data generating process: electricity demand is dictated by which appliances and systems are currently drawing power. Most appliances are used for contiguous stretches of time. For example, we turn a light on when we enter a room, and turn it off when we leave some time later. Further, many appliances have characteristic periods of use:  a microwave is typically on for a few minutes, while a washing machine is on for around an hour. A static model cannot capture these patterns.
	
	The held-out set average MSEs with bounds are shown in Table~\ref{table:real1}. The DDIBP performs comparably with the static model, suggesting its form of dependence is not appropriate for this task. The DIBP performs slightly better than the static model, indicating that it can capture the feature persistence described above. However, our model significantly outperforms the other models.  This can be explained by two properties of our model that are not present in the comparison methods. First, our method of modeling feature persistence is a natural fit for the data set: latent features are turned on at a rate given by the IBP, and they have an explicit duration that is independent of this rate. 
	
	By contrast, in the DIBP, a single Gaussian process controls both the rate at which a feature is turned on, and the amount of time for which it contributes. Second, our construction allows multiple instances of the same feature to contribute to a given time point. This means that our approach allows a single feature to model multiple similar appliances -- e.g. light bulbs -- which can be used simultaneously. The IFDM also performs favorably for this task, as we could imagine this problem to be like a blind signal separation problem where we want to model the probability when a dishwasher or laundry machine is on at a certain time point which is the setting for which such a model is designed. The IBP, DIBP and DDIBP, by contrast, must use separate features for different numbers of similar appliances in use, such as light bulbs.
	
	Consider a visual assessment of the importance of allowing multiple instances by examining the latent structures obtained from the static IBP  and our dynamic npLFM. Figures~\ref{fig:hpc_dynamic} and \ref{fig:hpc_ibp}, respectively, show the latent structure obtained from a single sample from these models. The top left panel of each of these figures shows the levels of the observed features. We can see that observed features 2 and 6 have spikes between observations 250 and 300. These spikes can be seen more clearly in Figure~\ref{fig:time_series} which plots the use of observed features 2 and 6 over time. Feature 2 corresponds to the minute-averaged voltage, and feature 6 corresponds to watt-hours of active energy in the third sub-circuit, which powers an electric water heater and an air-conditioner ---  both power-hungry appliances. The spikes are likely due to either the simultaneous use of the air-conditioner and water heater, or different levels of use of these appliances.

	Under the dynamic model, the bottom left panel of Figure~\ref{fig:hpc_dynamic} depicts that latent feature 0 places mass on observed features 2 and 6. The top right panel shows that there are multiple instances of this feature in observations corresponding to the previously discussed spikes in observed features 2 and 6. The corresponding static model graph in \ref{fig:hpc_ibp}  shows that the static IBP is unable to account for this latent behavior resulting from increased usage of the third sub-circuit;  hence this model must use a combination of multiple features to capture the same behavior.

	\subsection{Audio Real Data Illustration}\label{sec:audio1}
	It is natural to think of a musical piece in terms of latent features, for it is made up of one or more instruments playing one or more notes simultaneously. There is clearly persistence of features, making the longitudinal model described in Section~\ref{sec:model_long} a perfect fit. We chose to evaluate the model on a section of Strauss's ``Also Sprach Zarathrustra''. A midi-synthesized multi-instrumental recording of the piece was converted to a mono wave recording with an 8kHz sampling rate. We then generated a sequence of $D=128$-dimensional observations by applying a short-time Fourier transform using a  $128$-point discrete Fourier transform, a $128$-point Hanning window, and an 128-point advance between frames---so, each datum corresponds to a 16ms segment with a 16ms advance between segments. We scaled the data along each frequency component to have unit variance, and subtracted the minimum value for each observed feature.
	
	To evaluate the model, a hold-out sample of 10\% of the data, evenly spaced throughout the piece, was set aside. All but eight randomly selected dimensions were held out. Again, we use the same settings as described in the earlier experiments. We obtained average MSEs, along with bounds, by averaging over 5 independent trials from the final value of the Gibbs sampler, and are reported in Table~\ref{table:real1}. We see that by modeling the duration of features, we can perform favorably in a musical example which exhibits durational properties unlike the other models we compared against. Recall that the dynamic model has two advantages over the static model: it explicitly models feature duration, and allows multiple instances of a given feature to contribute to a given observation. The first aspect allows the model to capture the duration of notes or chords. The second allows the model to capture dynamic ranges of a single instrument, and the effect of multiple instruments playing the same note. 
	
	\subsection{Bird Call Source Separation}
	Next, we consider the problem of separating different sources of audio. Since it is difficult to imagine the number of different audio sources \textit{a priori}, we could instead learn the number of sources non-parametrically. A dynamic Indian buffet process model is well suited to this type of problem, as we may imagine different but possibly repeating sources of audio represented as the dishes selected from an IBP. To this end, we apply our dynamic model to an audio sample of bird calls. The audio source that we will look at for this problem is a two minute long recording of various bird calls in Kerala, India\footnote{The recording is available at \url{https://freesound.org/s/27334/}}. We transformed the raw wave file by Cholesky whitening the data and then took a regularly spaced subsample of 2,000 observations, of which we held out $10\%$ of the data randomly as a test set. We then analysed the data as described in Section~\ref{sec:audio1}.
	
	One could easily imagine that a bird call would be a natural duration element and would reappear throughout the recording. Hence, for a recording such as this one, being able to incorporate durational effects would be important to modeling the data. Though equivalently, one could also imagine this mode, again, like a blind source separation problem for which we could imagine a model like the IFDM performing favorably, without needing to model the durational component of the features. As seen in Table~\ref{table:real1}, we obtain superior performance for reasons, we posit, that we described above. 


	\section{Conclusion}\label{sec:conc}
	This paper introduces a new family of longitudinally dependent latent factor (or feature) models for time-dependent data. Unobserved latent features are often subject to temporal dynamics in data arising in a multitude of applications in industry. Static models for time-dependence exist but, as shown in this work, such approaches disregard key insights that could be gained if time dependency were to be modeled dynamically. Synthetic and real data illustrations exemplify the improved predictive accuracy while  using time-dependent, nonparametric latent feature models. General algorithms to sample from the new family developed here could be easily adapted to model data arising in different applications where the likelihood function changes. 
	
	This paper focused on temporal dynamics for random, \emph{fixed}, time-dependent data using nonparametric LFMs. But if data are \emph{changing} in real time, as in moving images in a film, then the notion of temporal dependency needs a different treatment than the one developed here. We wish to investigate this type of dependence in future work. In addition to the mathematical challenges that this proposed extension presents, the computational challenges are daunting as well. The theoretical and empirical work exhibited here show promise and we hope to develop faster and more expressive non-parametric factor models.

	\section*{Acknowledgments}
	Sinead Williamson and Michael Zhang were supported by NSF grant 1447721.
	
	
	
	\bibliographystyle{apalike}
	\bibliography{ibp}

\end{document}